# Tactical Generation in a Free Constituent Order Language

**Dilek Zeynep Hakkani, Kemal Oflazer, and Ilyas Cicekli**

Department of Computer Engineering and Information Science

Faculty of Engineering, Bilkent University, 06533 Bilkent, Ankara, Turkey

{hakkani,ko,ilyas}@cs.bilkent.edu.tr

## Abstract

This paper describes tactical generation in Turkish, a free constituent order language, in which the order of the constituents may change according to the information structure of the sentences to be generated. In the absence of any information regarding the information structure of a sentence (i.e., topic, focus, background, etc.), the constituents of the sentence obey a default order, but the order is almost freely changeable, depending on the constraints of the text flow or discourse. We have used a recursively structured finite state machine for handling the changes in constituent order, implemented as a right-linear grammar backbone. Our implementation environment is the GenKit system, developed at Carnegie Mellon University–Center for Machine Translation. Morphological realization has been implemented using an external morphological analysis/generation component which performs concrete morpheme selection and handles morphographemic processes.

## Introduction

Natural Language Generation is the operation of producing natural language sentences using specified communicative goals. This process consists of three main kinds of activities (McDonald, 1987):

- the goals the utterance is to obtain must be determined,

- the way the goals may be obtained must be planned,

- the plans should be realized as text.

Tactical generation is the realization, as linear text, of the contents specified usually using some kind of a feature structure that is generated by a higher level process such as text planning, or transfer in machine translation applications. In this process, a generation grammar and a generation lexicon are used.

As a component of a large scale project on natural language processing for Turkish, we have undertaken the development of a generator for Turkish sentences. In order to implement the variations in the constituent order dictated by various information structure constraints, we have used a recursively structured finite state machine instead of enumerating grammar rules for all possible word orders. A second reason for this approach is that many constituents, especially the arguments of verbs are typically optional and dealing with such optionality within rules proved to be rather problematic. Our implementation is based on the GenKit environment developed at Carnegie Mellon University–Center for Machine Translation. GenKit provides writing a context-free backbone grammar along with feature structure constraints on the non-terminals.

The paper is organized as follows: The next section presents relevant aspects of constituent order in Turkish sentences and factors that determine it. We then present an overview of the feature structures for representing the contents and the information structure of these sentences, along with the recursive finite state machine that generates the proper order required by the grammatical and information structure constraints. Later, we give the highlights of the generation grammar architecture along with some example rules and sample outputs. We then present a discussion comparing our approach with similar work, on Turkish generation and conclude with some final comments.

# Turkish

In terms of word order, Turkish can be characterized as a *subject–object–verb (SOV) language* in which constituents at certain phrase levels can change order rather freely, depending on the constraints of text flow or discourse. The morphology of Turkish enables morphological markings on the constituents to signal their grammatical roles without relying on their order. This, however, does not mean that word order is immaterial. Sentences with different word orders reflect different pragmatic conditions, in that, topic, focus and background information conveyed by such sentences differ.[1] Information conveyed through intonation, stress and/or clefting in fixed word order languages such as English, is expressed in Turkish by changing the order of the constituents. Obviously, there are certain constraints on constituent order, especially, inside noun and postpositional phrases. There are also certain constraints at sentence level when explicit case marking is not used (e.g., with indefinite direct objects).

In Turkish, the information which links the sentence to the previous context, the *topic*, is in the first position. The information which is new or emphasized, the *focus*, is in the immediately preverbal position, and the extra information which may be given to help the hearer understand the sentence, the *background*, is in the post verbal position (Erguvanlı, 1979). The topic, focus and background information, when available, alter the order of constituents of Turkish sentences. In the absence of any such control information, the constituents of Turkish sentences have the default order:

*subject, expression of time, expression of place, direct object, beneficiary, source, goal, location, instrument, value designator, path, duration, expression of manner, verb.*

All of these constituents except the verb are optional unless the verb obligatorily subcategorizes for a specific lexical item as an object in order to convey a certain (usually idiomatic) sense. The definiteness of the direct object adds a minor twist to the default order. If the direct object is an indefinite noun phrase, it has to be immediately preverbal. This is due to the fact that, both the subject and the indefinite direct object have no surface case-marking that distinguishes them, so word order constraints come into play to force this distinction.

In order to present the flavor of word order variations in Turkish, we provide the following examples. These two sentences are used to describe the same event (i.e., have the same logical form), but they are used in different discourse situations. The first sentence presents constituents in a neutral default order, while in the second sentence 'bugün' (today) is the topic and 'Ahmet' is the focus:[2]

(1)

a.

Ahmet bugün evden       okula
Ahmet today home+ABL school+DAT
'Ahmet went from home to school

otobüsle    3 dakikada  gitti.
bus+WITH 3 minute+LOC go+PAST+3SG
by bus in 3 minutes today.'

b.

Bugün evden      okula        otobüsle
today home+ABL school+DAT bus+WITH
'It was Ahmet who went from home to

3 dakikada    Ahmet gitti.
3 minute+LOC Ahmet go+PAST+3SG
school in 3 minutes by bus today.'

Although, sentences (b) and (c), in the following example, are both grammatical, (c) is not acceptable as a response to the question (a):

(2)

a.

Ali nereye      gitti?
Ali where+DAT go+PAST+3SG
'Where did Ali go?'

b.

Ali okula      gitti.
Ali school+DAT go+PAST+3SG
'Ali went to school.'

c.

\* Okula      Ali gitti.
school+DAT Ali go+PAST+3SG
'It was Ali who went to school.'

---

[1] See Erguvanlı (1979) for a discussion of the function of word order in Turkish grammar.

[2] In the glosses, 3SG denotes third person singular verbal agreement, P1PL and P3SG denote first person plural and third person singular possessive agreement, WITH denotes a derivational marker making adjectives from nouns, LOC, ABL, DAT, GEN denote locative, ablative, dative, and genitive case markers, PAST denotes past tense, and INF denotes a marker that derives an infinitive form from a verb.

The word order variations exemplified by (2) are very common in Turkish, especially in discourse.

## Generation of Free Word Order Sentences

The generation process gets as input a feature structure representing the content of the sentence where all the lexical choices have been made, then produces as output the surface form of the sentence. The feature structures for sentences are represented using a case-frame representation. Sentential arguments of verbs adhere to the same morphosyntactic constraints as the nominal arguments (e.g., the participle of, say, a clause that acts as a direct object is case-marked accusative, just as the nominal one would be). This enables a nice recursive embedding of case-frames of similar general structure to be used to represent sentential arguments.

In the next sections, we will highlight relevant aspects of our feature structures for sentences and their constituents.

### Simple Sentences

We use the case-frame feature structure in Figure 1 to encode the contents of a sentence.[3] We use the information given in the CONTROL feature to guide our grammar in generating the appropriate sentential constituent order. This information is exploited by a right linear grammar (recursively structured nevertheless) to generate the proper order of constituents at every sentential level (including embedded sentential clauses with their own information structure). The simplified outline of this right linear grammar is given as a finite state machine in Figure 2. Here, transitions are labeled by constraints and constituents (shown in bold face along a transition arc) which are generated when those constraints are satisfied. If any transition has a NIL label, then no surface form is generated for that transition.

The recursive behavior of this finite state machine comes from the fact that the individual argument or adjunct constituents can also embed sentential clauses. Sentential clauses correspond to either full sentences with non-finite or participle verb forms which act as noun phrases in either argument or adjunct roles, or gapped sentences with participle verb forms which function as modifiers of noun phrases (the filler of the gap). The former non-gapped forms can in Turkish be further classified into those representing *acts*, *facts* and *adverbials*. The latter (gapped form) is linked to the filler noun phrase by the ROLES feature in the structure for noun phrase (which will be presented in the following sections): this feature encodes the (semantic) role filled by the filler noun phrase and the case-frame of the sentential clause. The details of the feature structures for sentential clauses are very similar to the structure for the case-frame. Thus, when an argument or adjunct, which is a sentential clause, is to be realized, the clause is recursively generated by using the same set of transitions. For example, the verb 'gör' (see) takes a direct object which can be a sentential clause:

(3)

[3]Here, *c-name* denotes a feature structure for representing noun phrases or case-frames representing embedded sentential forms which can be used as nominal or adverbial constituents.

| S-FORM | infinitive/adverbial/participle/finite |
| CLAUSE-TYPE | existential/attributive/predicative |
| VOICE | active/reflexive/reciprocal/passive/causative |
| SPEECH-ACT | imperative/optative/necessitative/wish/ interrogative/declarative |

| QUES | TYPE | yes-no/wh |
| | CONST | list-of(subject/dir-obj/etc.) |

| VERB | ROOT | verb |
| | POLARITY | negative/positive |
| | TENSE | present/past/future |
| | ASPECT | progressive/habitual/etc. |
| | MODALITY | potentiality |

| ARGS | SUBJECT | c-name |
| | DIR-OBJ | c-name |
| | SOURCE | c-name |
| | GOAL | c-name |
| | LOCATION | c-name |
| | BENEFICIARY | c-name |
| | INSTRUMENT | c-name |
| | VALUE | c-name |

| ADJN | TIME | c-name |
| | PLACE | c-name |
| | MANNER | c-name |
| | PATH | c-name |
| | DURATION | c-name |

| CONTROL | TOPIC | constituent |
| | FOCUS | constituent |
| | BACKGR | constituent |

Figure 1: The case-frame for Turkish sentences.

Ayşe'nin    gelişini
Ayşe+GEN come+INF+P3SG
'I did not see Ayşe's coming.'

görmedim.
see+NEG+PAST+1SG

Similarly, the subject or any other constituent of a sentence can also be a sentential clause:

(4)

Ali'nin    buraya gelmesi
Ali+GEN here    come+INF+P3SG
'Ali's coming here made us

bizim    işi    bitirmemizi
we+GEN the job finish+INF+P1PL+ACC
finish the job easier.'

kolaylaştırdı.
make_easy+PAST+3SG

In all these cases, the main sentence generator also generates the sentential subjects and objects, in addition to generating the main sentence.

## Complex Sentences

Complex sentences are combinations of simple sentences (or complex sentences themselves) which are linked by either conjoining or various relationships like conditional dependence, cause–result, etc. The generator works on a feature structure representing a complex sentence which may be in one of the following forms:

- *a simple sentence.* In this case the sentence has the case-frame as its argument feature structure.

$$\begin{bmatrix} \text{TYPE} & \text{simple} \\ \text{ARG} & \text{case-frame} \end{bmatrix}$$

- *a series of simple or complex sentences* connected by coordinating or bracketing conjunctions. Such sentences have feature structures which have the individual case-frames as the values of their ELEMENTS features:

$$\begin{bmatrix} \text{TYPE} & \text{conj} \\ \text{CONJ} & \text{and/or/etc.} \\ \text{ELEMENTS} & \text{list-of(complex-sentence)} \end{bmatrix}$$

- *sentences linked with a certain relationship.* Such sentences have the feature structure:

$$\begin{bmatrix} \text{TYPE} & \text{linked} \\ \text{LINK-RELATION} & \text{rel} \\ \text{ARG1} & \text{complex-sentence} \\ \text{ARG2} & \text{complex-sentence} \end{bmatrix}$$

## Issues in Representing Noun Phrases

In this section we will briefly touch on relevant aspects of the representation of noun phrases. We use the following feature structure (simplified by leaving out irrelevant details) to describe the structure of a noun phrase:

$$\begin{bmatrix} \text{REF} & \begin{bmatrix} \text{ARG} & \textit{basic-concept} \\ \text{CONTROL} & \begin{bmatrix} \text{DROP} & +/- \text{ (default --)} \end{bmatrix} \end{bmatrix} \\ \text{CLASS} & \text{classifier} \\ \text{ROLES} & \textit{role-type} \\ \text{MODF} & \begin{bmatrix} \text{MOD-REL} & \text{list-of(}\textit{mod. relation}\text{)} \\ \text{ORDINAL} & \begin{bmatrix} \text{POSITION} & \textit{pos.} \\ \text{INTENSIFIER} & +/- \end{bmatrix} \\ \text{QUANT-MOD} & \textit{quantifier} \\ \text{QUALY-MOD} & \text{list-of(}\textit{simple-property}\text{)} \\ \text{CONTROL} & \begin{bmatrix} \text{EMPHASIS} & \text{quant./} \\ & \text{qual.} \end{bmatrix} \end{bmatrix} \\ \text{SPEC} & \begin{bmatrix} \text{DET} & \begin{bmatrix} \text{QUANTIFIER} & \textit{quant.} \\ \text{DEFINITE} & +/- \\ \text{REFERENTIAL} & +/- \\ \text{SPECIFIC} & +/- \end{bmatrix} \\ \text{SET-SPEC} & \text{list-of(}\textit{c-name}\text{)} \\ \text{SPEC-REL} & \text{list-of(}\textit{spec. relation}\text{)} \\ \text{DEMONS} & \textit{demonstrative} \end{bmatrix} \\ \text{POSS} & \begin{bmatrix} \text{ARGUMENT} & \textit{c-name} \\ \text{CONTROL} & \begin{bmatrix} \text{DROP} & +/- \\ \text{MOVE} & +/- \end{bmatrix} \end{bmatrix} \end{bmatrix}$$

The order of constituents in noun phrases is rather strict at a gross level, i.e., speficiers almost always precede modifiers and modifiers almost always precede classifiers,[4] which precede the head noun, although there are numerous exceptions. Also, within each group, word order variation is possible due to a number of reasons:

- The order of quantitative and qualitative modifiers may change: the aspect that is emphasized is closer to the head noun. The indefinite singular determiner may also follow

---

[4] A classifier in Turkish is a nominal modifier which forms a noun–noun noun phrase, essentially the equivalent of *book* in forms like *book cover* in English.

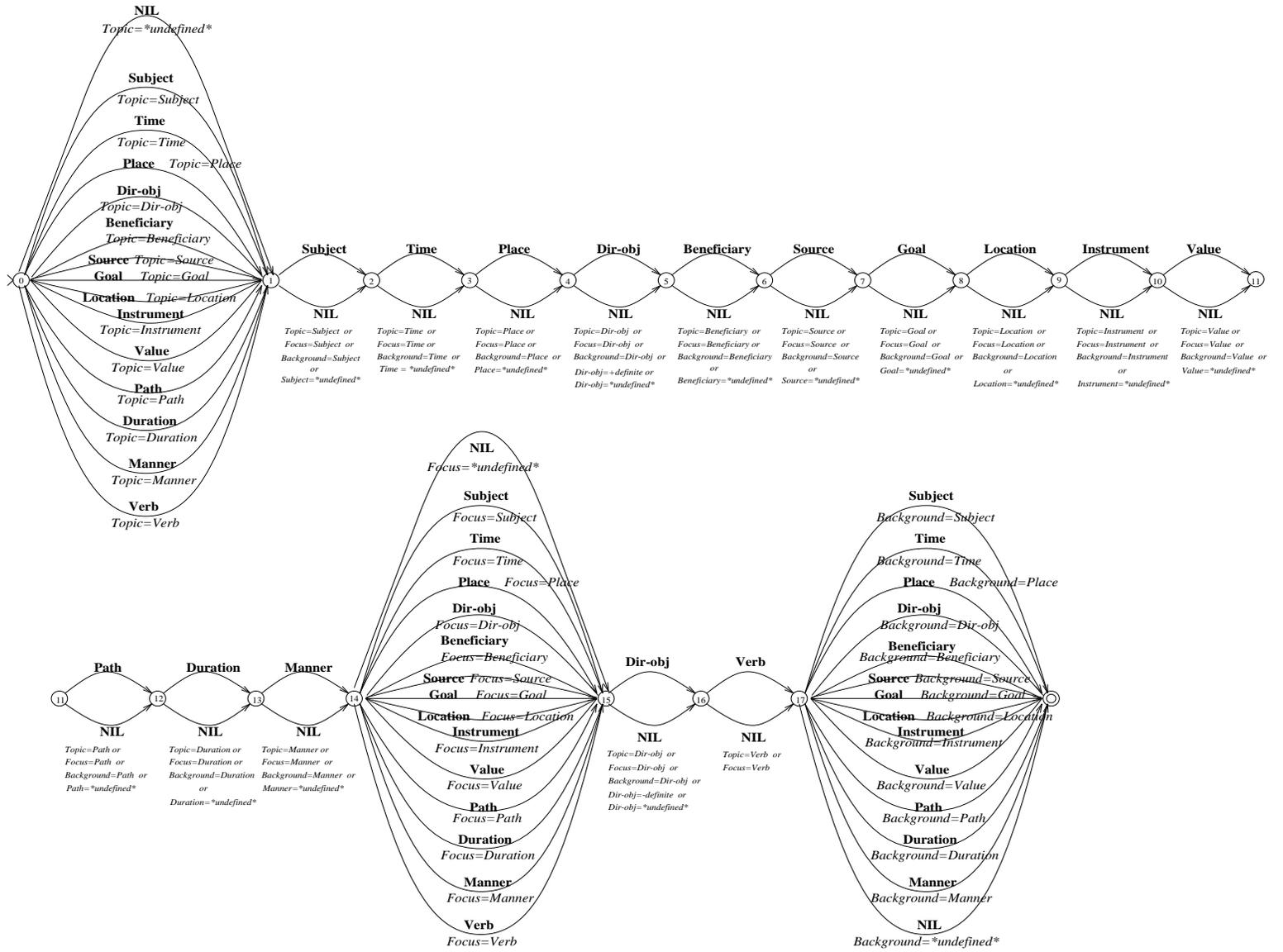

Figure 2: The finite state machine for generating the proper order of constituents in Turkish sentences.

any qualitative modifier and immediately precede any classifier and/or head noun.

- Depending on the determiner used, the position of the demonstrative specifier may be different. This is a strictly lexical issue and not explicitly controlled by the feature structure, but by the information (stored in the lexicon) about the determiner used.

- The order of lexical and phrasal modifiers (e.g., corresponding to a postpositional phrase on the surface) may change, if positioning the lexical modifier before the phrasal one causes unnecessary ambiguity (i.e., the lexical modifier in that case can also be interpreted as a modifier of some internal constituent of the phrasal modifier). So, phrasal modifiers always precede lexical modifiers and phrasal specifiers precede lexical specifiers, unless otherwise specified, in which case punctuation needs to be used.

- The possessor may scramble to a position past the head or even outside the phrase (to a background position), or allow some adverbial adjunct intervene between it and the rest of the noun phrase, *causing a discontinuous constituent*. Although we have included control information for scrambling the possessor to post head position, we have opted not to deal with either discontinuous constituents or long(er) distance scrambling as these are mainly used in spoken discourse.

Furthermore, since the possessor information is explicitly marked on the head noun, if the discourse does not require an overt possessor[5] it may be dropped by suitable setting of the DROP feature.

## Interfacing with Morphology

As Turkish has complex agglutinative word forms with productive inflectional and derivational morphological processes, we handle morphology outside our system using the generation component of a full-scale morphological

---

[5]For example, (c) cannot be used as an answer to (a) in the following discourse, where the owner of the book should be emphasized:

a.  Kimin      kitabı       kalın?
    whose      book+P3SG    thick
    'Whose book is thick?'

b.  Benim      kitabım      kalın.
    I+GEN      book+P1SG    thick
    'My book is thick.'

c.  * Kitabım          kalın.
      book+P1SG        thick

analyzer of Turkish (Oflazer, 1993). Within GenKit, we generate relevant abstract morphological features such as agreement and possessive markers and case marker for nominals and voice, polarity, tense, aspect, mood and agreement markers for verbal forms. This information is properly ordered at the interface and sent to the morphological generator, which then:

1. performs concrete morpheme selection, dictated by the morphotactic constraints and morphophonological context,

2. handles morphographemic phenomena such as vowel harmony, and vowel and consonant ellipsis, and

3. produces an agglutinative surface form.

## Grammar Architecture and Output

Our generation grammar is written in a formalism called Pseudo Unification Grammar implemented by the GenKit generation system (Tomita and Nyberg, 1988). Each rule consists of a context-free phrase structure description and a set of *feature constraint equations*, which are used to express constraints on feature values. Non-terminals in the phrase structure part of a rule are referenced as $x0,\ldots,xn$ in the equations, where $x0$ corresponds to the non-terminal in the left hand side, and $xn$ is the $n^{th}$ non-terminal in the right hand side. Since the context-free rules are directly compiled into tables, the performance of the system is essentially independent of the number of rules, but depends on the complexity of the feature constraint equations (which are compiled into LISP code). Currently, our grammar has 273 rules each with very simple constraint checks. Of these 273 rules, 133 are for sentences and 107 are for noun phrases.

To implement the sentence level generator (described by the finite state machine presented earlier), we use rules of the form

$$S_i \rightarrow XP\ S_j$$

where the $S_i$ and $S_j$ denote some state in the finite state machine and the $XP$ denotes the constituent to be realized while taking this transition. If this $XP$ corresponds to a sentential clause, the same set of rules are recursively applied. This is a variation of the method suggested by Takeda *et al.* (1991).

The following are rule examples that implement some of the transitions from state 0 to state 1:

```
(<S> <==> (<S1>)
    (
     ((x0 control topic) =c *undefined*)
     (x1 = x0)
    ))

(<S> <==> (<Subject> <S1>)
    (
     ((x0 control topic) =c subject)
     (x2 = x0)
     ((x2 arguments subject) = *remove*)
     (x1 = (x0 arguments subject))
    ))

(<S> <==> (<Time> <S1>)
    (
     ((x0 control topic) =c time)
     (x2 = x0)
     ((x2 adjuncts time) = *remove*)
     (x1 = (x0 adjuncts time))
    ))
```

The grammar also has rules for realizing a constituent like `<Subject>` or `<Time>` (which may eventually call the same rules if the argument is sentential) and rules like above for traversing the finite state machine from state 1 on.

## Examples

In this section, we provide feature structures for three example sentences which only differ in their information structures. Although the following feature structures seem very similar, they correspond to different surface forms.[6]

(5)

Ahmet dün kitabı masada
Ahmet yesterday book+ACC table+LOC
'Ahmet left the book on the table

bıraktı.
leave+PAST+3SG
yesterday.'

---

$$
\begin{bmatrix}
\text{S-FORM} & \text{finite} \\
\text{CLAUSE-TYPE} & \text{predicative} \\
\text{VOICE} & \text{active} \\
\text{SPEECH-ACT} & \text{declarative} \\
\text{VERB} & \begin{bmatrix} \text{ROOT} & \text{\#bırak} \\ \text{SENSE} & \text{positive} \\ \text{TENSE} & \text{past} \\ \text{ASPECT} & \text{perfect} \end{bmatrix} \\
\text{ARGUMENTS} & \begin{bmatrix} \text{SUBJECT} & \{\text{Ahmet}\} \\ \text{DIR-OBJ} & \{\text{kitap}\} \\ \text{LOCATION} & \{\text{masa}\} \end{bmatrix} \\
\text{ADJUNCTS} & \begin{bmatrix} \text{TIME} & \{\text{dün}\} \end{bmatrix}
\end{bmatrix}
$$

(6)

Dün kitabı masada Ahmet
yesterday book+ACC table+LOC Ahmet
'It was Ahmet who left the book on

bıraktı.
leave+PAST+3SG
the table yesterday.'

$$
\begin{bmatrix}
\text{S-FORM} & \text{finite} \\
\text{CLAUSE-TYPE} & \text{predicative} \\
\text{VOICE} & \text{active} \\
\text{SPEECH-ACT} & \text{declarative} \\
\text{VERB} & \begin{bmatrix} \text{ROOT} & \text{\#bırak} \\ \text{SENSE} & \text{positive} \\ \text{TENSE} & \text{past} \\ \text{ASPECT} & \text{perfect} \end{bmatrix} \\
\text{ARGUMENTS} & \begin{bmatrix} \text{SUBJECT} & \{\text{Ahmet}\} \\ \text{DIR-OBJ} & \{\text{kitap}\} \\ \text{LOCATION} & \{\text{masa}\} \end{bmatrix} \\
\text{ADJUNCTS} & \begin{bmatrix} \text{TIME} & \{\text{dün}\} \end{bmatrix} \\
\text{CONTROL} & \begin{bmatrix} \text{TOPIC} & \text{time} \\ \text{FOCUS} & \text{subject} \end{bmatrix}
\end{bmatrix}
$$

(7)

Dün kitabı Ahmet
yesterday book+ACC Ahmet
'It was Ahmet who left the book

bıraktı masada.
leave+PAST+3SG table+LOC
yesterday on the table.'

| S-FORM | finite |
| CLAUSE-TYPE | predicative |
| VOICE | active |
| SPEECH-ACT | declarative |

$$
\begin{array}{ll}
\text{VERB} & \begin{bmatrix} \text{ROOT} & \text{\#bırak} \\ \text{SENSE} & \text{positive} \\ \text{TENSE} & \text{past} \\ \text{ASPECT} & \text{perfect} \end{bmatrix} \\[2em]
\text{ARGUMENTS} & \begin{bmatrix} \text{SUBJECT} & \{\text{Ahmet}\} \\ \text{DIR-OBJ} & \{\text{kitap}\} \\ \text{LOCATION} & \{\text{masa}\} \end{bmatrix} \\[2em]
\text{ADJUNCTS} & \begin{bmatrix} \text{TIME} & \{\text{dün}\} \end{bmatrix} \\[1em]
\text{CONTROL} & \begin{bmatrix} \text{TOPIC} & \text{time} \\ \text{FOCUS} & \text{subject} \\ \text{BACKGROUND} & \text{location} \end{bmatrix}
\end{array}
$$

Figure 3 shows the path the generator follows while generating sentence 7. The solid lines show the transitions that the generator makes in its right linear backbone.

## Comparison with Related Work

Dick (1993) has worked on a classification based language generator for Turkish. His goal was to generate Turkish sentences of varying complexity, from input semantic representations in Penman's Sentence Planning Language (SPL). However, his generator is not complete, in that, noun phrase structures in their entirety, postpositional phrases, word order variations, and many morphological phenomena are not implemented. Our generator differs from his in various aspects: We use a case-frame based input representation which we feel is more suitable for languages with free constituent order. Our coverage of the grammar is substantially higher than the coverage presented in his thesis and we also use a full-scale external morphological generator to deal with complex morphological phenomena of agglutinative lexical forms of Turkish, which he has attempted embedding into the sentence generator itself.

Hoffman, in her thesis (Hoffman, 1995a, Hoffman, 1995b), has used the Multiset–Combinatory Categorial Grammar formalism (Hoffman, 1992), an extension of Combinatory Categorial Grammar to handle free word order languages, to develop a generator for Turkish. Her generator also uses relevant features of the information structure of the input and can handle word order variations within embedded clauses. She can also deal with scrambling out of a clause dictated by information structure constraints, as her formalism allows this in a very convenient manner. The word order information is lexically kept as multisets associated with each verb. She has demonstrated the capabilities of her system as a component of a prototype database query system. We have been influenced by her approach to incorporate information structure in generation, but, since our aim is to build a wide-coverage generator for Turkish for use in a machine translation application, we have opted to use a simpler formalism and a very robust implementation environment.

## Conclusions

We have presented the highlights of our work on tactical generation in Turkish – a free constituent order language with agglutinative word structures. In addition to the content information, our generator takes as input the information structure of the sentence (topic, focus and background) and uses these to select the appropriate word order. Our grammar uses a right-linear rule backbone which implements a (recursive) finite state machine for dealing with alternative word orders. We have also provided for constituent order and stylistic variations within noun phrases based on certain emphasis and formality features. We plan to use this generator in a prototype transfer-based human assisted machine translation system from English to Turkish.

## Acknowledgments

We would like to thank Carnegie Mellon University–Center for Machine Translation for providing us the GenKit environment. This work was supported by a NATO Science for Stability Project Grant TU-LANGUAGE.

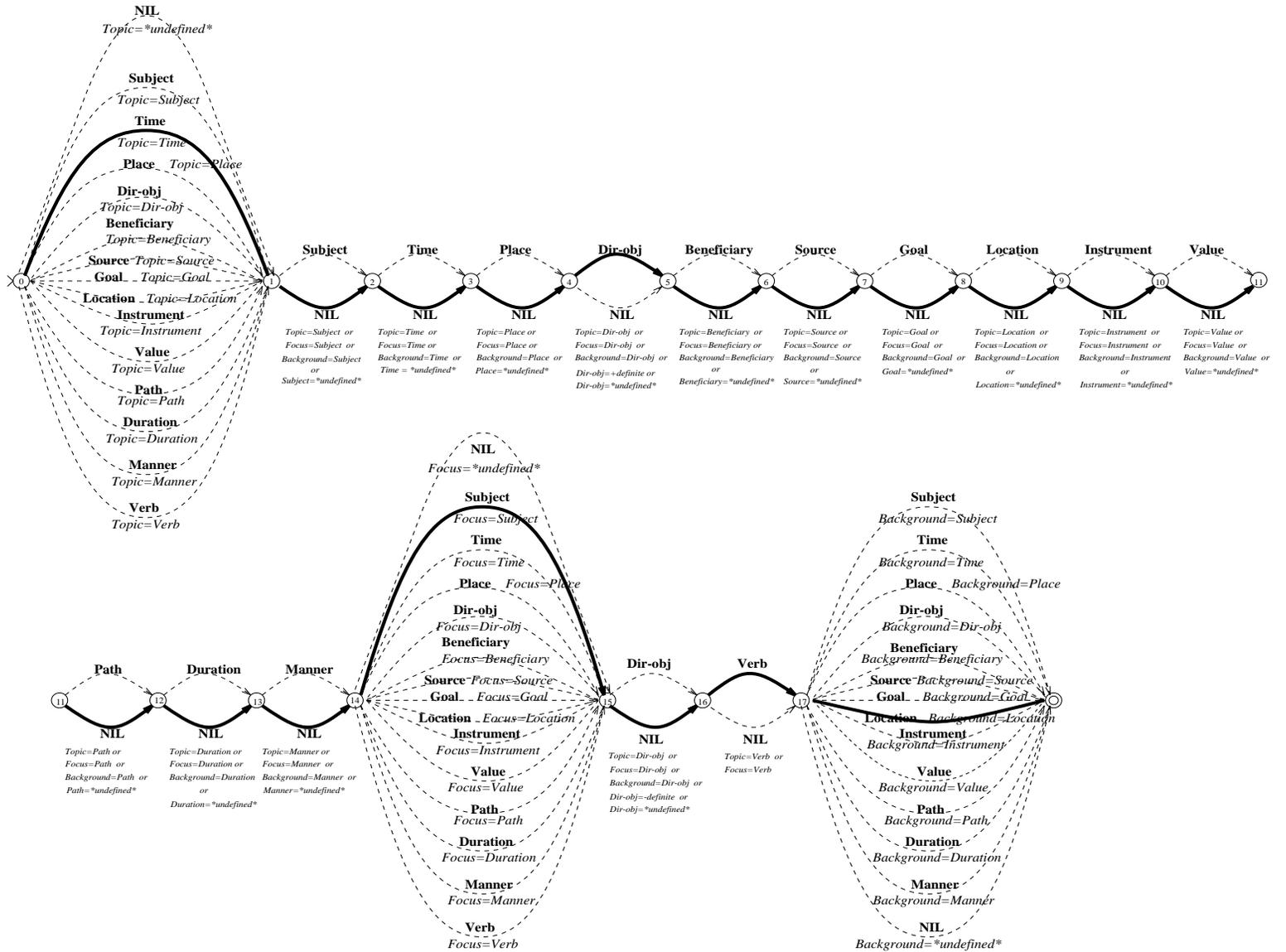

Figure 3: The transitions followed for generating sentence 7.